# Deep Learning Discovery of Demographic Biomarkers in Echocardiography

Grant Duffy[1], Shoa L. Clarke[2], Matthew Christensen[1], Bryan He[3], Neal Yuan[4], Susan Cheng[1], and David Ouyang[1,5]

1. Department of Cardiology, Smidt Heart Institute, Cedars-Sinai Medical Center
2. Department of Medicine, Division of Cardiovascular Medicine, Stanford University
3. Department of Computer Science, Stanford University
4. San Francisco Veteran Affairs Medical Center, University of California San Francisco
5. Division of Artificial Intelligence in Medicine, Department of Medicine, Cedars-Sinai Medical Center

Email: David.Ouyang@cshs.org

**Abstract**

Deep learning has been shown to accurately assess 'hidden' phenotypes and predict biomarkers from medical imaging beyond traditional clinician interpretation of medical imaging. Given the black box nature of artificial intelligence (AI) models, caution should be exercised in applying models to healthcare as prediction tasks might be short-cut by differences in demographics across disease and patient populations. Using large echocardiography datasets from two healthcare systems, we test whether it is possible to predict age, race, and sex from cardiac ultrasound images using deep learning algorithms and assess the impact of varying confounding variables. Using a total of 433,469 cardiac ultrasound videos from Cedars-Sinai Medical Center and 99,909 videos from Stanford Medical Center, we trained video-based convolutional neural networks to predict age, sex, and race. We found that deep learning models were able to identify age and sex, while unable to reliably predict race. Without considering confounding differences between categories, the AI model predicted sex with an AUC of 0.85 (95% CI 0.84 – 0.86), age with a mean absolute error of 9.12 years (95% CI 9.00 – 9.25), and race with AUCs ranging from 0.63 - 0.71. When predicting race, we show that tuning the proportion of a confounding variable (sex) in the training data significantly impacts model AUC (ranging from 0.57 to 0.84), while in training a sex prediction model, tuning a confounder (race) did not substantially change AUC (0.81 – 0.83). This suggests a significant proportion of the model's performance on predicting race could come from confounding features being detected by AI. Further work remains to identify the particular imaging features that associate with demographic information and to better understand the risks of demographic identification in medical AI as it pertains to potentially perpetuating bias and disparities.

# Introduction

Recent advances in deep learning have resulted in leaps in performance in analyzing and assessing image data, both with natural images as well as diagnostic medical images[1–3]. While traditional computer vision datasets are used by algorithms to perform common perceptive visual tasks that are achievable by most humans (for example, recognizing a cat when an animal is in the image)[2,4,5], deep learning in medicine has extended to tasks of prognosis and detection beyond the normal abilities of human clinicians. From evaluating blood pressure in fundoscopic images[6] to predicting prognosis and biomarkers from videos[7,8], convolutional neural networks are being applied to tasks not traditionally performed by clinicians.

Recent work from a variety of researchers on many medical imaging modalities have suggested deep learning can identify demographic features from medical waveforms, images, or videos[6,8–12]. The ability of deep learning algorithms to identify age, sex, and race from medical imaging raises challenging questions about whether using artificial intelligence (AI) black box models can be a vector for perpetuating biases and disparities [13–15]. The current regulatory environment does not require the demonstration of standard performance across different populations[16], however it has been shown that even when race is not directly used as an input, complicated decision support systems can learn patterns that reinforce disparities in access or treatment[17].

In this analysis, we sought to systematically evaluate whether demographic information can be captured from echocardiography, cardiac ultrasound, videos using deep learning. Using video-based deep learning architectures known to be able to capture quantitative traits commonly assessed by clinicians as well as textual patterns associated with disease[18,19], we evaluate whether echocardiogram videos can be used to predict age, sex, and race, and whether these models are robust across varying confounding variables and generalize across institutions with different geographic and demographic characteristics.

## Results

**Study Cohort Characteristics**

This study used cohorts of patients from two geographically distinct independent health care systems with different patient demographics. The Cedars-Sinai Medical Center (CSMC) study cohort consists of 30,762 patients who underwent 51,640 echocardiogram studies between 2011 and 2021. The mean age at the time of echocardiogram study was 66.5±16.4 years, 44.8% were women, and 68.8% self-identified as White. The SHC study cohort consists of 99,909 patients who underwent 99,909 echocardiogram studies between 2000 and 2019. The mean age at the time of echocardiogram study was 59.9±17.7 years, 44.3% were women, and 56.5% self-identified as White. Demographic characteristics are shown in **Table 1**.

When initially trained on 99,909 apical 4 chamber videos from SHC without balancing confounding covariables, video-based deep learning models successfully learned features of age, sex, and race. The deep learning model accurately predicted sex with AUC of 0.93 on the hold-out test set and 0.85 on the external test set of apical 4 chamber views from CSMC. Predicting Black, Asian, and White races, the AI model achieved an AUC of 0.74, 0.73, and 0.71 respectively on the hold-out test set and 0.71, 0.66, and 0.63 on the external test set. The AI model predicted age with a MAE of 7.40 years on the hold-out test set and 9.29 years on the external test set. In general, there was a slight drop off in model performance with external validation as shown in **Figure 1**, however the model learned generalizable features that allowed for high accuracy in external validation datasets with sufficient training examples. When trained on echocardiogram videos from CSMC without balancing confounding covariables, video-based deep learning models similarly successfully learned features of age and sex and, to a lesser extent, race. With 150,913 CSMC apical 4 chamber video training examples, the deep learning model accurately predicted sex with an AUC of 0.84, age with a MAE of 9.66 years, and race with an AUC ranging from 0.54 – 0.60 on the held out test dataset.

We hypothesized that the modest ability to predict race from echocardiograms may depend on biased distributions of predictable features in race-stratified cohorts. For example, in the CSMC data, the proportion of males among the White subset is slightly higher (58.5%) compared to the proportion of males in the Black subset (54.1%). To test the impact of bias in a single predictable covariate, we artificially created subset datasets where race was confounded by sex. When sex is matched (bias = 0.5) in the training set, the model performance decreased for predicting White race (AUC 0.57) compared to the model trained without matching (AUC 0.59). Further, as sex bias was increased, the model performance approaches that of the sex prediction task (AUC 0.84) suggesting that the model was primarily identifying image features consistent with sex (**Figure 2**). When trained to predict sex on these datasets, the performance remained constant regardless of how biased the data was to race. Thus, race adds little to no signal when predicting sex with a biased dataset (AUC ranging from 0.81 to 0.83). These results demonstrate how even a single confounder may be used by a deep learning model to predict race.

We expect that in most real-world datasets of clinical data, there are many relevant confounders that vary across race categories. Some of these confounders are comorbidities as shown in **Table 2**. To evaluate the predictive power of these confounders, we used logistic regression to predict White from non-White from comorbidity data alone. This method resulted in an AUC of 0.62, comparable to the AI computer vision model trained on CSMC data. Many of these cardiovascular comorbidities have known cardiac imaging findings. If an AI is able to detect imaging findings of these confounders, then it would be able to "shortcut" predict race without learning any additional information.

**Discussion**

In this study, we systemically evaluated whether deep learning models can learn features of age, sex, and race from large datasets of echocardiogram videos. Consistent with prior applications of AI to medical imaging, we show that echocardiogram videos can be used to accurately predict age and sex and the models generalize across institutions. In contrast, we were not able to reproduce similarly accurate predictions of race and the model performance did not generalize well across institutions. Our experiments suggest race prediction results from shortcutting through predictions of known confounders commonly seen when stratifying large population cohorts by race.

Consistent with our deep learning model's high accuracy in predicting age and sex, there are well known age-associated changes [20,21] and sexual dimorphism[22,23] in cardiac structures visualized by ultrasound. While clinicians do not routinely use echocardiography to assess age, established age-dependent references for echocardiography and cardiac MRI highlights the recognized changes seen with normal cardiac remodeling of aging[24]. Conversely, conventionally measured metrics in echocardiography do not significantly vary across race, often within measurement error across cohorts[21,24]. Others have shown that AI models can predict race accurately using chest X-rays, thoracic CT scans, and breast mammograms[12], but we were not able to replicate such results in echocardiography. Such results, if not interpreted thoughtfully, risk sending the dangerous and false message that race reflects distinct biological features. This falsehood has been the basis of several missteps in medicine[25]. Thus, it is critical that AI researchers understand and acknowledge how and why deep learning models can predict both biologic (e.g. sex) and non-biologic (e.g. race) features.

Our results suggest that AI models may predict race by leveraging the non-random distribution of predictable features in race-stratified cohorts. We use sex, a single highly predictable feature, to demonstrate that the degree of bias in a predictable confounding feature can arbitrarily impact prediction of a downstream task. Consequentially, we expect that there are many imaging-relevant features with varying predictability that contribute to the performance of AI race prediction from medical imaging. Importantly, some features, particularly related to health and disease, reflect the impacts of social determinant of health, and thus systemic differences in these features by racial category in population cohorts is a marker of the structural racism in medicine and society[26]. Failure to recognize these nuances in medical applications of AI, and the misapplication of population cohorts without evaluating confounding variables, could lead to repeating mistakes of the past[25].

There are a few limitations in this study worth mentioning. First, the echocardiogram videos used in this study were only of two institutions, although geographically distinct and with different population demographics. The predominant ultrasound machine make and model was the same in both institution, which could standardize input video information and facilitate external validity but through imaging characteristics that are particular to that particular ultrasound machine. Second, the medical imaging used in this study were obtained in the course of routine clinical care, thus are enriched for individuals with access to healthcare and comorbidities that might have particular relationships with age, sex, and race. While this work was motivated to understand if AI models can shortcut prediction tasks through predicting demographics, there are biases regarding who is able to access healthcare, at what stage of disease, and for whom imaging is obtained. Ethical considerations must be considered carefully, as fair application of AI is required to avoid perpetuating or exacerbating current biases in the healthcare system.

In summary, echocardiogram videos contain information that is detectable by AI models and is predictive of demographic features. Age and sex features appear to be recognizable and generalize across geography and institution, while race prediction has significant dependence on the construction of the training dataset and performance could be mostly attributable to bias in training data. Further work remains to identify the particular imaging features that associate with demographic information and to better understand the risks of demographic identification in medical AI as it pertains to potentially perpetuating bias and disparities.

## Methods

### Datasets

We used echocardiogram video data from two large academic medical centers, Cedars-Sinai Medical Center (CSMC) and Stanford Healthcare (SHC). Originally stored as DICOM videos after acquisition by GE or Phillips ultrasound machines, we used a standard pre-processing workflow to remove information outside of the ultrasound sector, identify metadata[27], and save files in AVI format. Videos were stored as 112x112 pixel video files and view classified into four standard echocardiographic views (apical-4-chamber, apical-2-chamber, parasternal long axis, and subcostal views).

Echocardiogram videos were split into training, validation, and test datasets by patient to prevent data leakage across splits. Demographic information was obtained from the electronic health record. Age was calculated from time from the echocardiogram study to date of birth. Information about sex and race were obtained from the electronic health record based on self-report from the clinical record. Comorbidities were extracted from the electronic health record by International Classification of Disease Nineth or Tenth Revision codes present in problem lists or visit associated diagnoses within 1 year of the echocardiogram imaging study. This analysis did not independently re-survey or use other instruments to evaluate data labels.

For experiments sweeping model performance with various degrees of biased training datasets, we subsetted the whole dataset to form simplified cohorts that binarized demographic categories and maintained the same training set size across experiments. The breakdown of these subgroups can be found in **Tables 3 and 4**. For predicting race, we focused on white/non-white binary classification and varying the confounding variable of sex. For example, a bias of 0.5 corresponds to a dataset where 50% of both white and non-white examples are male and female. A bias of 1.0 corresponds to a dataset where all of the white patients are male and all of the non-white patients are female. Other than training dataset construction, all other model training details (architectures, loss, learning rate, number of epochs, etc) were held the same. The parallel set of experiments predicting sex used the same format only with race proportion in the training set varied as the cofounding variable.

**AI Model Architecture**

Spatiotemporal relationships were captured by our deep learning model using 3D convolutions using standard ResNet architecture (R2+1D)[18]. Models were trained to minimize L2 loss for predicting age (a regression task), binary cross entropy for predicting sex (a binary classification task), and cross entropy for predicting race (a multi-class classification task). Models were trained using stochastic gradient descent with an ADAM optimizer using an initial learning rate of 0.01, momentum of 0.9, and learning rate decay. Models were trained using an array of NVIDIA 2080, 3090, and A6000 graphical processing units.

**Analysis**

The performance of deep learning models was assessed on internal held out datasets or external datasets from another institution. The performance of predicting age was evaluated by the mean absolute difference between the model prediction and actual age at time of echocardiogram study. The prediction of sex and race was evaluated by area under receiver operating curve (AUROC). Confidence intervals were computed using 10,000 bootstrapped samples of the test datasets. To benchmark with the predictive ability of differences in population level comorbidity rates, we developed a logistic regression model using all comorbidities in Table 2 as well as age and sex as independent input variables to predict race. The continuous output of the logistic regression model was used to assess AUROC. This research was approved by the Stanford University and Cedars-Sinai Medical Center Institutional Review Boards.

**Table 1.** Demographic characteristics of study participants

|  | CSMC | | | | SHC |
|---|---|---|---|---|---|
|  | Apical 4 Chamber | Apical 2 Chamber | Parasternal Long Axis | Subcostal | Apical 4 Chamber |
| n, patients | 28,450 | 25,502 | 28,685 | 23,596 | 99,909 |
| n, videos | 186,426 | 71,086 | 110,399 | 65,558 | 99,909 |
| Age (mean (SD)) | 66.5 (±16.5) | 66.7 (±16.5) | 66.1 (±16.5) | 66.2 (±16.4) | 59.9 (±17.7) |
| Male (%) | 15,713 (55.2%) | 14,093 (55.3%) | 15,739 (54.9%) | 12,884 (54.6%) | 55,610 (55.7%) |
| Race/Ethnicity, n (%) | | | | | |
|   American Indian | 65 (0.2%) | 57 (0.2%) | 66 (0.2%) | 56 (0.2%) | 267 (0.3%) |
|   Asian | 2,162 (7.6%) | 1,945 (7.6%) | 2,157 (7.5%) | 1,808 (7.7%) | 14,197 (14.2%) |
|   Black | 4,058 (14.3%) | 3,681 (14.4%) | 4,156 (14.5%) | 3,322 (14.1%) | 4,826 (4.8%) |
|   Pacific Islander | 87 (0.3%) | 82 (0.3%) | 86 (0.3%) | 75 (0.3%) | 1,428 (1.4%) |
|   White | 19,519 (68.6%) | 17,444 (68.4%) | 19,595 (68.3%) | 16,211 (68.7%) | 56,498 (56.5%) |
|   Other | 1,980 (7.0%) | 1,790 (7.0%) | 2,021 (7.0%) | 1,659 (7.0%) | 17,452 (17.5%) |
|   Unknown | 579 (2.0%) | 503 (2.0%) | 604 (2.1%) | 465 (2.0%) | 5,241 (5.2%) |

CSMC = Cedars-Sinai Medical Center, SHC = Stanford Healthcare

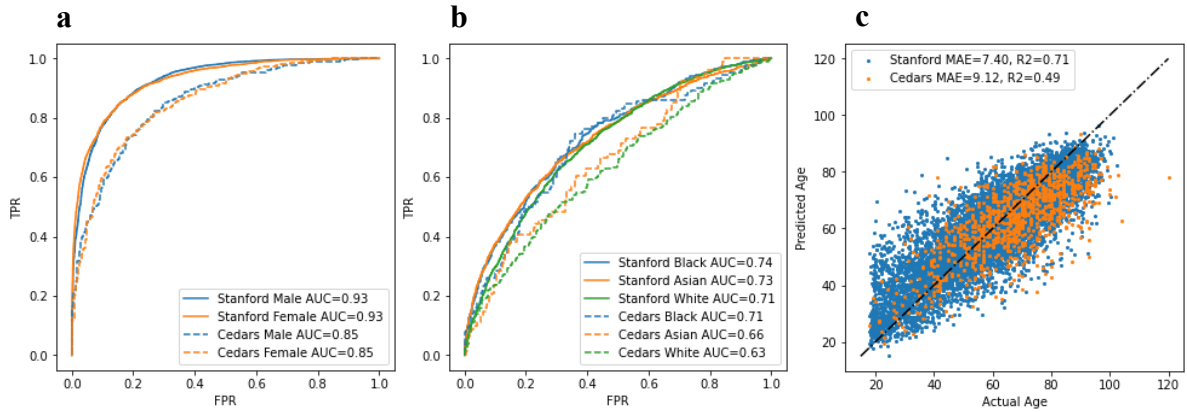

**Fig. 1.** Model performance for (a) sex, (b) race and (c) age prediction when trained on SHC apical-4-chamber echocardiogram videos and evaluated on SHC hold-out test set and CSMC external validation dataset unadjusted for confounders. CSMC = Cedars-Sinai Medical Center, SHC = Stanford Healthcare

**Fig. 2.** Model performance after varying training set balance of confounding variable. **a**, Model performance in predicting race when varying sex of training set. **b**, Model performance in predicting sex when varying race of training set. **c**, Summary table of results.

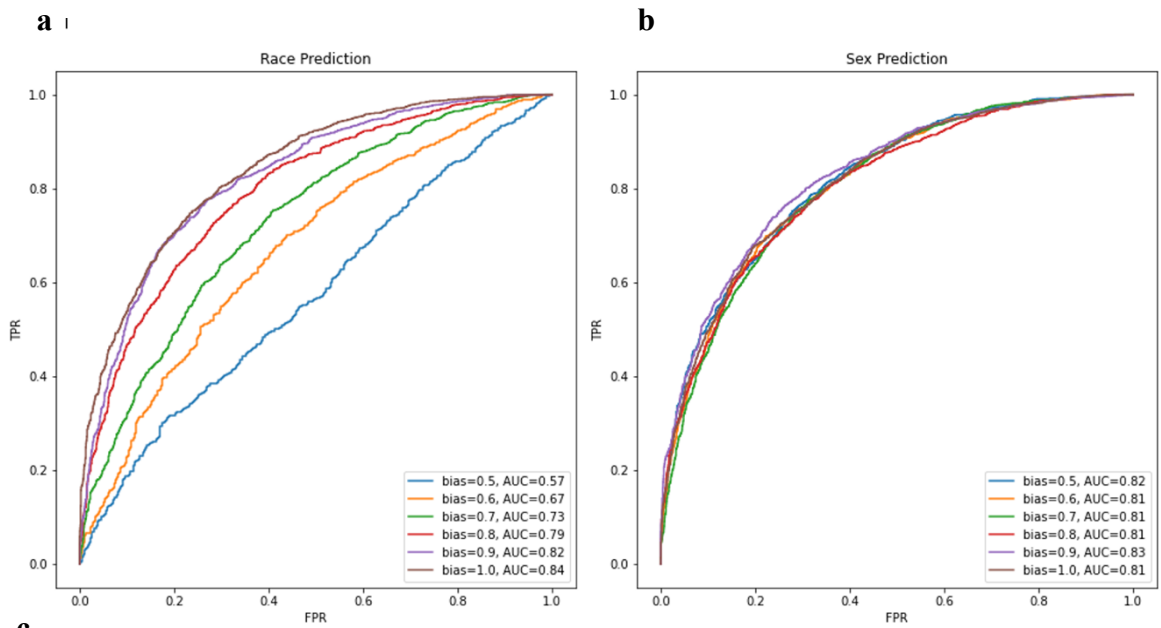

| Training Task To Predict Race (White/Not White) | | Training Task To Predict Sex (Male/ Not Male) | |
| --- | --- | --- | --- |
| Proportion Biased | AUC | Proportion Biased | AUC |
| 50% Male in White Cohort / Female in Non White | 0.57 | 50% White in Male Cohort / Non White in Not Male | 0.82 |
| 60% Male in White Cohort / Female in Non White | 0.67 | 60% White in Male Cohort / Non White in Not Male | 0.81 |
| 70% Male in White Cohort / Female in Non White | 0.73 | 70% White in Male Cohort / Non White in Not Male | 0.81 |
| 80% Male in White Cohort / Female in Non White | 0.79 | 80% White in Male Cohort / Non White in Not Male | 0.81 |
| 90% Male in White Cohort / Female in Non White | 0.82 | 90% White in Male Cohort / Non White in Not Male | 0.83 |
| 100% Male in White Cohort / Female in Non White | 0.84 | 100% White in Male Cohort / Non White in Not Male | 0.81 |

**Table 2.** Race subgroup comorbidities for Cedars-Sinai Medical Center cohort

| Prevalent diagnoses at time of study | White | Black or African American | Asian | Total |
|---|---|---|---|---|
| Atrial Fibrillation | 33,883 (29.83%) | 6,132 (24.83%) | 3,577 (27.46%) | 43,592 (28.81%) |
| Heart Failure | 49,411 (43.50%) | 13,084 (52.99%) | 5,990 (45.98%) | 68,485 (45.26%) |
| Hypertension | 68,114 (59.96%) | 17,561 (71.12%) | 8,014 (61.51%) | 93,689 (61.92%) |
| Diabetes | 26,511 (23.34%) | 8,292 (33.58%) | 4,392 (33.71%) | 39,195 (25.90%) |
| Ischemic Stroke | 12,017 (10.58%) | 4,190 (16.97%) | 1,405 (10.78%) | 17,612 (11.64%) |
| Transient Ischemic Attack | 7,431 (6.54%) | 1,457 (5.90%) | 410 (3.15%) | 9,298 (6.14%) |
| Systemic Embolism | 1,076 (0.95%) | 346 (1.40%) | 113 (0.87%) | 1,535 (1.01%) |
| Pulmonary Embolism | 4,981 (4.38%) | 2,146 (8.69%) | 340 (2.61%) | 7,467 (4.93%) |
| Prior Myocardial Infarction | 14,573 (12.83%) | 4,148 (16.80%) | 1,930 (14.81%) | 20,651 (13.65%) |
| Stroke/Transient Ischemic Attack/Thromboembolism | 32,387 (28.51%) | 9,380 (37.99%) | 3,529 (27.09%) | 45,296 (29.94%) |
| Peripheral Arterial Disease | 18,318 (16.13%) | 3,471 (14.06%) | 1,861 (14.28%) | 23,650 (15.63%) |
| Vascular Disease | 29,029 (25.56%) | 6,721 (27.22%) | 3,421 (26.26%) | 39,171 (25.89%) |
| Coronary Artery Disease | 40,848 (35.96%) | 6,977 (28.26%) | 4,664 (35.80%) | 52,489 (34.69%) |
| Chronic Kidney Disease | 28,346 (24.95%) | 10,148 (41.10%) | 4,044 (31.04%) | 42,538 (28.11%) |
| Liver Disease | 7,566 (6.66%) | 1,177 (4.77%) | 1,011 (7.76%) | 9,754 (6.45%) |
| Chronic Obstructive Pulmonary Disease | 7,329 (6.45%) | 2,474 (10.02%) | 387 (2.97%) | 10,190 (6.73%) |
| Prior Smoker | 7,077 (6.23%) | 2,089 (8.46%) | 678 (5.20%) | 9,844 (6.51%) |

**Table 3.** Race subgroup characteristics for Cedars-Sinai Medical Center cohort

| CSMS | White | Black or African American | Asian | Total |
|---|---|---|---|---|
| By Study | | | | |
|     Male n (%) | 73,925 (58.5%) | 14,791 (54.1%) | 8,005 (55.1%) | 96,721 (57.49%) |
|     Female n (%) | 52,468 (41.5%) | 12,527 (45.9%) | 6,536 (44.9%) | 71,531 (42.51%) |
|     Study Age mean (std) | 67.68 (16.4) | 63.0 (15.8) | 65.1 (16.6) | 66.69 (16.43) |
| Total | 126,393 | 27,318 | 14,541 | 168,252 |
| By Patient | | | | |
|     Male n (%) | 11,036 (56.5%) | 1,978 (48.7%) | 1,140 (52.7%) | 14,154 (54.99%) |
|     Female n (%) | 8,483 (43.5%) | 2,080 (51.3%) | 1,022 (47.3%) | 11,585 (45.01%) |
| Total | 19,519 | 4,058 | 2,162 | 25,739 |

**Table 4.** Race subgroup characteristics for Stanford Healthcare cohort

| SHC | White | Black or African American | Asian | Total |
|---|---|---|---|---|
| By Study | | | | |
|   Male n (%) | 32,150 (57.04%) | 2,490 (51.72%) | 7,397 (52.17%) | 42,037 (55,78%) |
|   Female n (%) | 24,217 (42.96%) | 2,324 (48.28%) | 6,781 (47.83%) | 33,322 (44.22%) |
|   Study Age mean (std) | 61.85 (17.37) | 56.21 (16.49) | 60.07 (17.05) | 61.15 (17.31%) |
| Total | 56,367 | 14,178 | 4,814 | 75,359 |